\documentclass[sn-mathphys]{sn-jnl}

\jyear{2021}%

\theoremstyle{thmstyleone}%
\theoremstyle{thmstyletwo}%

\theoremstyle{thmstylethree}%

\begin{document}
	\newcommand{\bfx}{{\mathbf{x}}}
	\newcommand{\bfh}{{\mathbf{h}}}
	\newcommand{\bfw}{{\mathbf{w}}}
	\newcommand{\bfv}{{\mathbf{v}}}
	
	\newcommand{\bfX}{{\mathbf{X}}}
	
	\newcommand{\calC}{{\mathcal{C}}}
	\newcommand{\calO}{{\mathcal{O}}}
	\newcommand{\calM}{{\mathcal{M}}}
	
	\newcommand{\calR}{{\mathcal{R}}}
	\newcommand{\calL}{{\mathcal{L}}}
	\newcommand{\calI}{{\mathcal{I}}}
	
	\newcommand{\calK}{{\mathcal{K}}}
	\newcommand{\calD}{{\mathcal{D}}}
	\newcommand{\calP}{{\mathcal{P}}}

	\title[MrTF: Model Refinery for Transductive Federated Learning]{MrTF: Model Refinery for Transductive Federated Learning}
	
	
	\author[1,2]{\fnm{Xin-Chun} \sur{Li}}\email{lixc@lamda.nju.edu.cn}

	\author*[3,4]{\fnm{Yang} \sur{Yang}}\email{yyang@njust.edu.cn}
	
	\author[1,2]{\fnm{De-Chuan} \sur{Zhan}}\email{zhandc@nju.edu.cn}
	
	\affil[1]{\orgdiv{State Key Laboratory for Novel Software Technology}, \orgname{Nanjing University}, \orgaddress{\city{Nanjing}, \postcode{210023}, \state{Jiangsu}, \country{China}}}
	
	\affil[2]{\orgdiv{School of Artificial Intelligence}, \orgname{Nanjing University}, \orgaddress{\city{Nanjing}, \postcode{210023}, \state{Jiangsu}, \country{China}}}
	
	\affil[3]{\orgdiv{School of Computer Science and Engineering}, \orgname{Nanjing University of Science and Technology}, \orgaddress{\city{Nanjing}, \postcode{210094}, \state{Jiangsu}, \country{China}}}
	
	\affil[4]{\orgdiv{Department of Computing}, \orgname{The Hong Kong Polytechnic University}, \orgaddress{\city{Hong Kong}, \postcode{100872}, \country{China}}}
	

	
	\abstract{We consider a real-world scenario in which a newly-established pilot project needs to make inferences for newly-collected data with the help of other parties under privacy protection policies. Current federated learning (FL) paradigms are devoted to solving the data heterogeneity problem without considering the to-be-inferred data. We propose a novel learning paradigm named transductive federated learning (TFL) to simultaneously consider the structural information of the to-be-inferred data. On the one hand, the server could use the pre-available test samples to refine the aggregated models for robust model fusion, which tackles the data heterogeneity problem in FL. On the other hand, the refinery process incorporates test samples into training and could generate better predictions in a transductive manner. We propose several techniques including stabilized teachers, rectified distillation, and clustered label refinery to facilitate the model refinery process. Abundant experimental studies verify the superiorities of the proposed \underline{M}odel \underline{r}efinery framework for \underline{T}ransductive \underline{F}ederated learning (MrTF). The source code is available at \url{https://github.com/lxcnju/MrTF}.}

	
	
	
	\maketitle
	
	\section{Introduction}
\begin{figure}[tbp]
		\includegraphics[width=\linewidth]{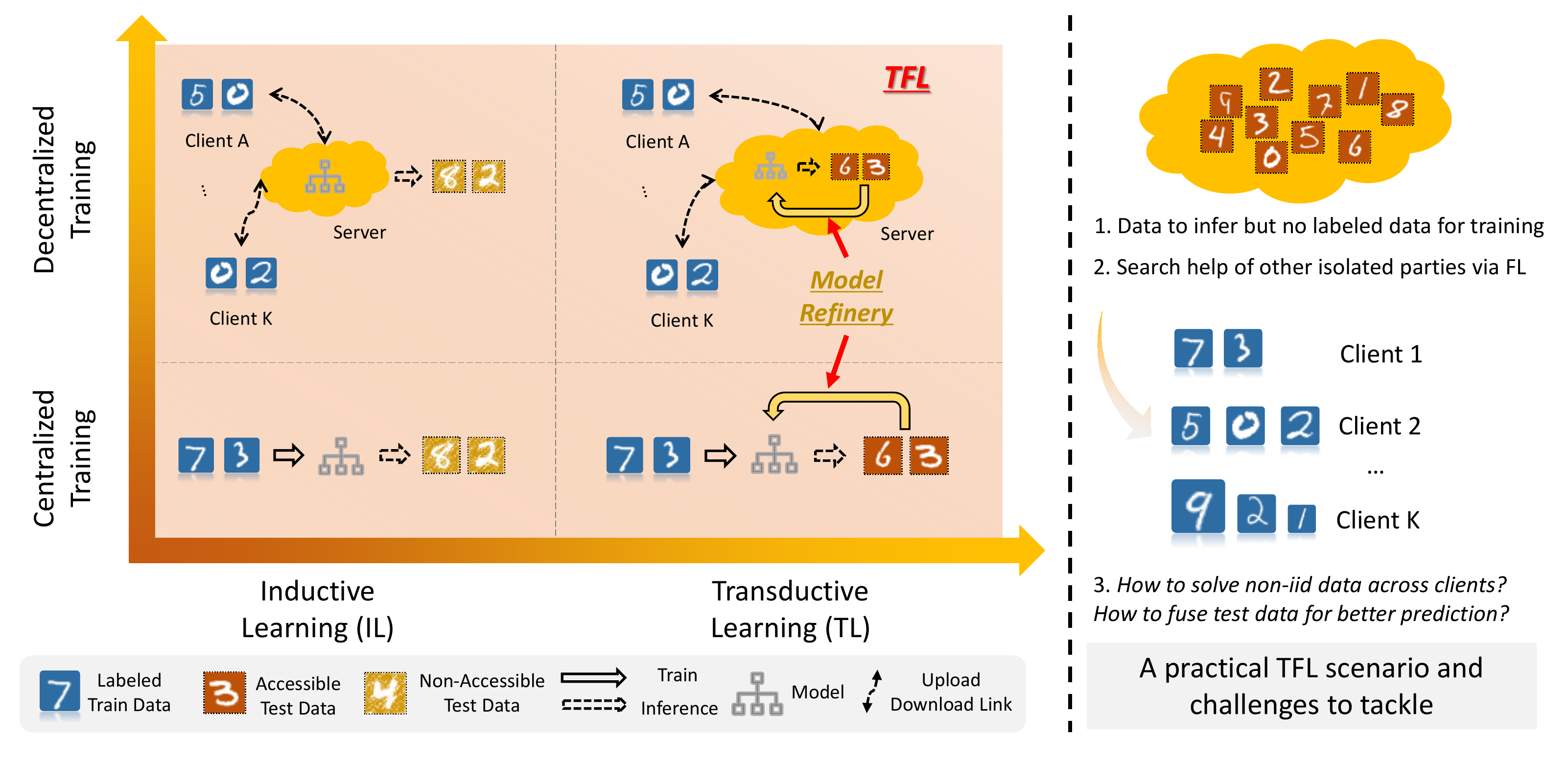}
		\caption{\small Left: comparisons of several paradigms. TFL could access test data in advance and the training data is decentralized across clients. Right: a real-world scenario formulated as TFL and corresponding challenges.}
		\label{fig:teaser}
	\end{figure}

	Machine learning techniques, especially deep learning, have been widely applied to various real-world applications~\cite{ResNet,Transformer,FedKWS-UI,TKDE-Inc}, etc. Inductive learning and transductive learning are two common learning paradigms, where the latter could obtain test samples in advance~\cite{TransductiveSVM,TransferAsTransductive,TransductiveFewShot-ICLR2019}. Fusing to-be-inferred unlabeled data into training could lead to appreciable performances because it simultaneously captures structural information in both train and test data. The settings of these two paradigms are illustrated at the bottom of Fig.~\ref{fig:teaser}.
	
	
	A practical scenario is that a particular party/user is urgent to make predictions for newly-collected unlabeled data while it has no labeled samples for training. Thus, it needs to seek the help of other relevant parties/users (clients) to build a prediction model in collaboration. There is a restriction that the data of other parties cannot be used directly due to privacy protection policies.
	Federated Learning (FL)~\cite{Fed-Concept,FedAvg,FedPAN} has been proposed as an efficient distributed training paradigm to collaborate with isolated parties without sending users' data out. On the one hand, various challenges have emerged in FL, e.g., the Non-Independent Identically Distributed (Non-I.I.D.) data challenge~\cite{NonIID-Quag}. Participating clients may own various data distributions under different contexts, leading to weight divergence in distributed local training~\cite{Fed-NonIID-Data}. On the other hand, existing FL follows an inductive manner that aims to build a global model that could generalize well to any possible forthcoming samples without considering the information of the pre-available test samples. The scenario and challenges are illustrated at the right of Fig.~\ref{fig:teaser}, where the newly-established party has 10 classes to infer, while the participating parties may contain only several classes, various classes, or imbalanced classes, etc. These challenges lead to Non-I.I.D. data that hinders the effectiveness of FL.
	
	In this paper, we abstract the scene mentioned above as {\it transductive federated learning (TFL), where the server owns to-be-inferred data in advance while training data are distributed across clients in a Non-I.I.D. manner}. In TFL, the goal of the server is to assign labels to the test samples on hand. There are two fundamental challenges to tackle: (1) {\it how to overcome the Non-I.I.D. challenge across clients during distributed training?} (2) {\it how to improve the inference process for unlabeled data on the server?} To solve the former, some previous FL works take various techniques. For example, some works~\cite{Fed-Shared-Data} allow sending out a small portion of clients' data. A better way to meet the privacy policy is the recently proposed FedDF~\cite{FedDF}, which views the local updated models as ``teachers" and distills the knowledge to the aggregated model on a publicly available dataset for robust model fusion. However, collecting appropriate public data is also challenging, especially in data-scarce scenarios. For the latter, we should consider the structural information contained in the pre-available test data to assign better predictions. Considering both, FedDF seems to be an appropriate solution. First, the trouble of collecting unlabeled public data is omitted in TFL because the to-be-referred data is a good candidate. What's more, distilling on the to-be-referred samples also considers their information, which is expected beneficial for making predictions.
	
	Nevertheless, it is not all smooth sailing. FedDF encounters several fatal challenges faced with large amounts of clients and stochastic client selection. Specifically, FedDF takes the ``AvgLogi" (i.e., Averaging Logits) of local updated models as the ensemble, while the logits' magnitudes across local models vary a lot, and directly averaging them as the ``teacher" may lead to training instability. Additionally, stochastic client selection does not guarantee that the local model's ensemble covers all classes, negatively transferring knowledge for missing classes. Correspondingly, we propose a more stable way as an alternative to refine the aggregated model via rectifying the local models' logits and introducing label clustering techniques. Verified on several benchmarks, our proposed methods show superiorities towards other methods. Our contributions could be briefed as originally introducing the practical TFL framework and proposing an effective solution named MrTF.
	
	
	
	\section{Related Works}
	Our work is closely related to federated learning (FL), transductive learning (TL), and external data in FL.

	\noindent \textbf{Federated Learning (FL).} FL~\cite{Fed-Concept,FedAvg} aims to organize isolated clients to accomplish the machine learning process following a distributed training style. As the most standard FL algorithm, Federated Averaging (FedAvg)~\cite{FedAvg} follows the parameter server architecture~\cite{ParameterServer} where a server coordinates amounts of clients. During the whole process, only model parameters are transmitted, and advanced privacy protection methods (e.g., differential privacy~\cite{DeepDP}) could be additionally applied for stricter privacy protections. The Non-I.I.D. challenge in FL refers to that data of participating users are heterogeneous under various contexts, which hinders the aggregation and personalization in FL~\cite{Fed-NonIID-Data,FedRS}. Various solutions are proposed for better model fusion via introducing local regularization~\cite{FedProx,FedMMD}, reducing gradient variance~\cite{Scaffold}, fine-tuning aggregated model via additional data~\cite{Fed-Shared-Data,FedDF}, etc. Knowledge distillation~\cite{KD} could facilitate the generalization in FL~\cite{FedAgnosticKD,FedDFKD}. However, these FL methods are only devoted to solving the data heterogeneity problem and could not be directly applied to simultaneously solve the two major challenges of TFL.

	\noindent \textbf{Transductive Learning (TL).} Inductive learning assumes test data are not available during training, requiring the trained models to generalize well on any possible test set. As the opposition, TL could access the to-be-inferred data, and the training process could progressively capture the structural information in both train and test data. TL relaxes the requirement of model generalization and only aims to make better predictions on the available test data. Hence, compared with inductive learning, TL could basically achieve better results when the test samples are accessible. TSVM~\cite{TransductiveSVM} utilizes the margin information in test samples and yields better SVM models. A real-world scenario is that we have to make predictions for unlabeled samples in a novel domain, and fusing labeled samples from source domains for together training is a common solution in transfer learning~\cite{TransferAsTransductive} or domain adaptation~\cite{DAN}. Another advantageous scene for TL is learning with few-shot samples, where some studies~\cite{TransductiveFewShot-ICLR2019} have verified the superiorities of TL.

	\noindent \textbf{External data in FL.} To reduce the weight divergence in FL, ~\cite{Fed-Shared-Data} utilize additional labeled data on the server to fine-tune the global model, while~\cite{FedMD} resorts to publicly available labeled data. Some semi-supervised FL also introduces unlabeled data~\cite{FedMatch,FedConSSL}, while they consider the clients own both labeled and unlabeled samples. The most related work to ours is FedDF~\cite{FedDF}, which utilizes ``AvgLogi" to ensemble local models and further distill the knowledge from them to the aggregated model. However, FedDF only considers the cross-silo scenes defined in~\cite{Fed-Advances} where the amount of local clients is small (e.g., 20 clients on CIFAR~\cite{cifar}) and the client participation ratio is high (e.g., 40\% on vision tasks and 100\% on NLP tasks). With large amounts of local clients and stochastic client selection, FedDF faces several problems. FedED~\cite{FedED} extends FedDF for medical relation extraction. More practical scenes of utilizing FedDF are also studied, e.g., the resource-aware scenes~\cite{ResourceFL}.
	
	\noindent \textbf{Other related works.} Learning from multiple source domains~\cite{MSTL-CVPR2010} is also related to TFL. The former does not consider the privacy protection policies and could send out source data or source models to facilitate the learning process of the target domain. The fundamental problem in these works is how to measure the transferability~\cite{MSTL-NeurIPS2021,Explore-ICASSP2022} between source domains and the target domain. TFL considers data privacy protection, making the learning process more challenging. In TFL, we aim to simultaneously tackle the data heterogeneity problem and make better predictions for the to-be-inferred data.
	
	
	\section{Preliminaries}
	In this section, we first detail the setting and goal of TFL. Then, we introduce FedAvg~\cite{FedAvg}/FedDF~\cite{FedDF} and their drawbacks in TFL.
	\subsection{Transductive Federated Learning (TFL)} \label{sec:tfl}
	TFL also follows the parameter server~\cite{ParameterServer} architecture, and assumes training data are decentralized on local clients while the server could previously access the to-be-inferred data. Mathematically, we have $K$ clients and each client owns a unique data distribution $\calD_k=p_{k}(\bfx,y)=p_k(\bfx)p_k(y\vert \bfx), k \in [K]$. We denote the observed samples as $\{(\bfx_{k,i}, y_{k,i})\}_{i=1}^{n_k}$, where $n_k$ is the number of training samples on $k$th client. The total number of training samples from all clients is $N=\sum_{k=1}^K n_k$. In TFL, we assume the server owns an unlabeled set $\{\bfx_j\}_{j=1}^M \sim p_{\text{g}}(\bfx)$ with $M$ samples to be predicted. The goal of TFL is to make good predictions on the test set via collaborating with these $K$ clients without transmitting clients' data. {\it Generally, we consider the data distribution of the test data (i.e., $p_{\text{g}}(\cdot)$) does not diverge a lot from the data distribution if all clients' data are centralized (i.e., $\frac{1}{K}\sum_{k=1}^K p_k(\cdot)$). We also consider the opposite case in Sect.~\ref{sec:cross-tfl} (i.e., cross-domain TFL).}
	
	\subsection{Federated Averaging (FedAvg)} \label{sec:fedavg}
	FedAvg~\cite{FedAvg} takes $T$ communication rounds of local and global procedures to collaborate with local clients. During local procedures, a small fraction (i.e. $R \in [0, 1]$) of clients $S_t$ download the global model from server and update it on their own local data for $E$ epochs. We denote the global model parameters in $t$th round as $\theta_t$, and the updated model on $k$th client is $\theta_{t,k}$. During the global procedure, the server collects the updated models and takes a simple parameter averaging process as $\theta_{t+1} \leftarrow \frac{1}{\vert S_t\vert } \sum_{k\in S_t} \theta_{t,k}$. Faced with heterogeneous data, the local model update incurs large gradient variance and weight divergence~\cite{Fed-NonIID-Data}. In the following, we denote $f_k(\bfx;\theta_k)$ or $f_k$ as the prediction function of the $k$th local model that outputs the ``logits" for $C$ classes, while $f_{\text{g}}(\bfx;\theta)$ or $f_{\text{g}}$ as the prediction function of the aggregated model. We sometimes omit the communication round index $t$ for simplification. We use $q_k(y\vert \bfx;\theta_k) = \sigma(f_k(\bfx;\theta_k))$ to denote the predicted class probability distribution based on the $k$th local model, where $\sigma(\cdot)$ is the softmax operator. Similarly, $q_{\text{g}}(y\vert \bfx;\theta)$ denotes the predicted probability of the aggregated model. Notably, $q(\cdot;\theta)$ denotes predicted probabilities while $p(\cdot)$ denotes the oracle ones.
	
	\subsection{Non-I.I.D. Data} \label{sec:Non-I.I.D.}
\begin{figure}[tbp]
		\centering
		
		\begin{minipage}{0.24\linewidth}
			\centering
			\includegraphics[width=\linewidth]{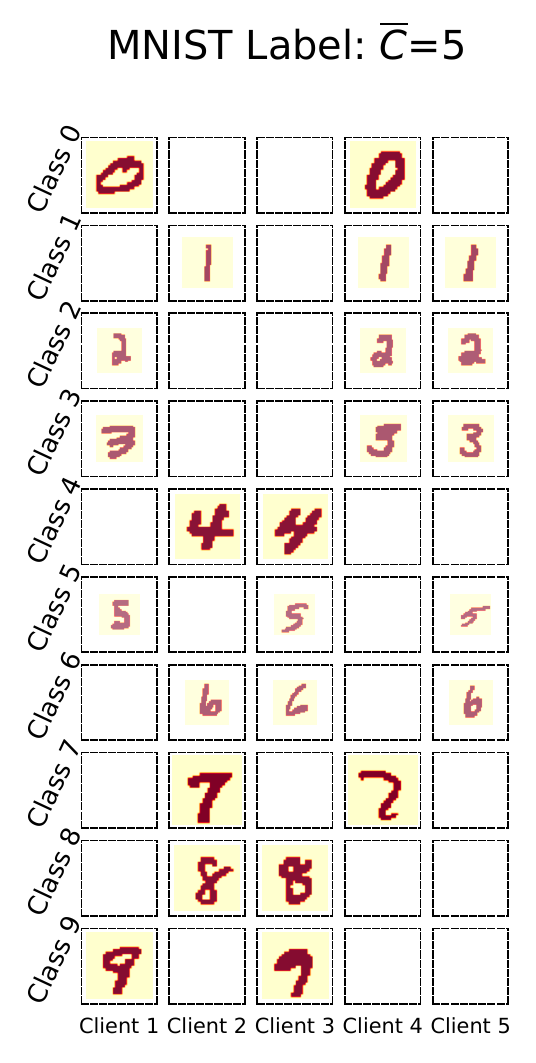}
		\end{minipage}
		\begin{minipage}{0.24\linewidth}
			\centering
			\includegraphics[width=\linewidth]{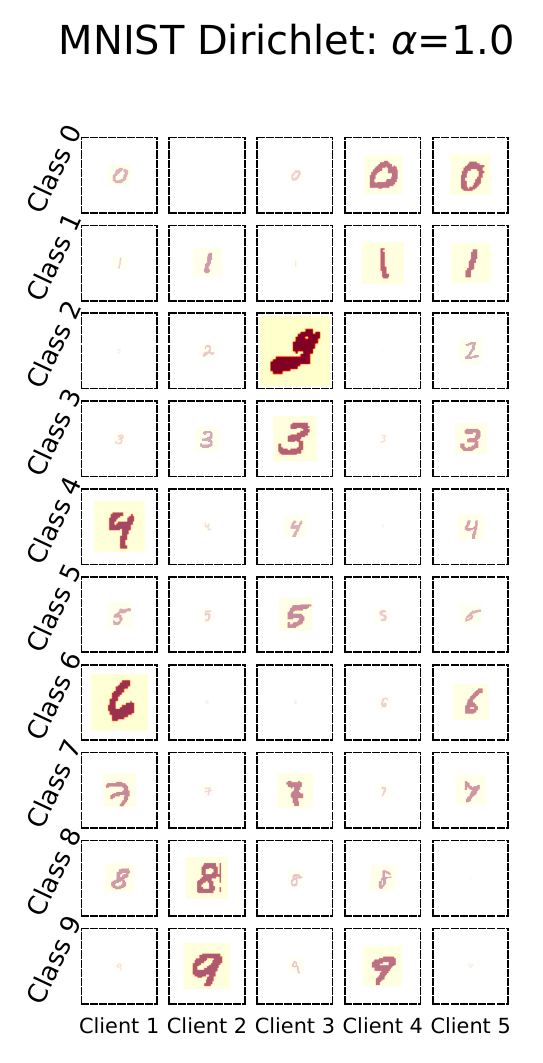}
		\end{minipage}
		\begin{minipage}{0.24\linewidth}
			\centering
			\includegraphics[width=\linewidth]{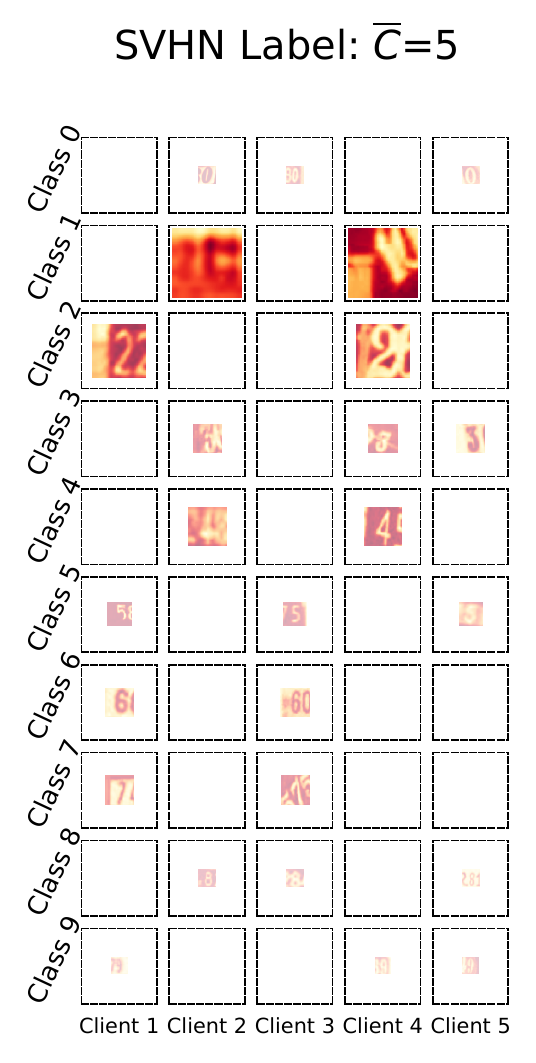}
		\end{minipage}
		\begin{minipage}{0.24\linewidth}
			\centering
			\includegraphics[width=\linewidth]{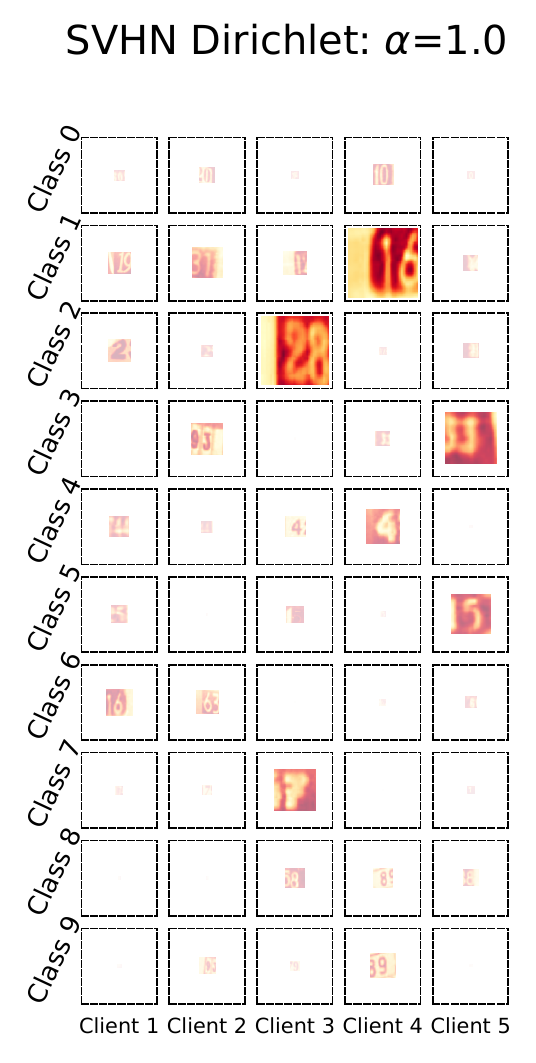}
		\end{minipage}
		
		\centering
		\caption{\small Split data distributions with $K=5$ clients. We split MNIST and SVHN via both ``split by label" ($\overline{C} = 5$) and ``split by dirichlet" ($\alpha = 1.0$). Darker colors and larger sizes mean more samples.}
		\label{fig:data}
	\end{figure}

	Because users' data is generated from different contexts, the data across federated clients is usually Non-I.I.D.. For experimental studies, previous works distribute a definite public data set (e.g., MNIST~\cite{mnist}, CIFAR~\cite{cifar}) onto $K$ clients according to various split strategies. In classification tasks with $C$ classes, two commonly utilized ways are ``{\it split by label}" and ``{\it split by dirichlet}". The former assumes each client could only observe $\overline{C}$ classes while other $C-\overline{C}$ classes are not accessible~\cite{FedAvg,Fed-NonIID-Data,FedRS}. Although some classes are missing, the observed classes are almost balanced. A smaller $\overline{C}$ corresponds to more serious Non-I.I.D. data. The latter samples a class distribution from the Dirichlet distribution $p_k(y) \sim \text{Dir}(\alpha)$ for each client~\cite{NonIID-Quag,FedDF}, where $\alpha$ controls the Non-I.I.D. level, and a smaller $\alpha$ corresponds to a more Non-I.I.D. scene. After determining local clients' class distributions, training data are accordingly allocated to these clients for distributed training. We study both cases in this paper and show the split distributions with $K=5$ clients in Fig.~\ref{fig:data}. {\it These two cases cover both challenges caused by data heterogeneity, data imbalance, amount skew, and missing classes, which are sufficient to verify the effectiveness of proposed FL methods.} Aside from these constructed Non-I.I.D. scenes split by classes, we also consider benchmarks split by users in experimental studies.
	
\begin{figure*}[tbp]
	\centering
	
	\begin{minipage}{\linewidth}
		\centering
		\includegraphics[width=\linewidth]{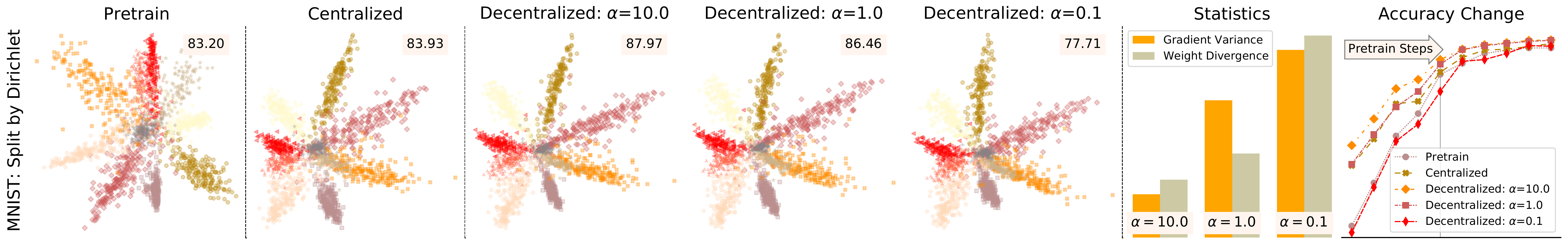}
	\end{minipage}
	\quad
	\begin{minipage}{\linewidth}
		\centering
		\includegraphics[width=\linewidth]{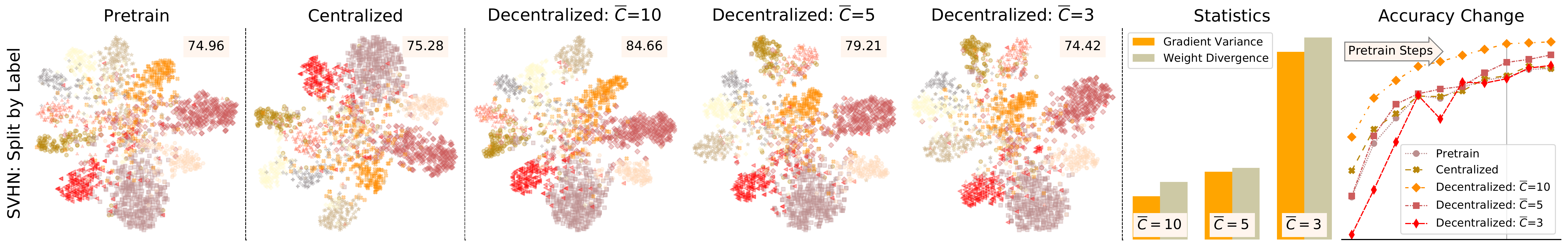}
	\end{minipage}
	
	\centering
	\caption{\small Performance degradation of FedAvg under Non-I.I.D. scenes. The two rows show ``split by dirichlet" on MNIST and ``split by label" on SVHN, respectively. In each row, the left five figures show the extracted features and test accuracies (top-right numbers) of the pre-trained model, centralized model, and decentralized model under three levels of Non-I.I.D. data. The bars show two measures to evaluate the divergence of distributed training and centralized training. The rightmost shows the accuracy change of these five models with respect to pre-training steps.}
	\label{fig:obser}
	\end{figure*}

	We implement FedAvg on several Non-I.I.D. cases via a one-shot FL similar to~\cite{OneShotFL} with only one communication round. Specifically, we first pre-train a global model on the centralized training set for $r_0$ steps and denote the obtained parameters as $\theta_0$. We use $\theta_0$ as initializations for both centralized training and decentralized training. For the former, we continually update $\theta_0$ on the centralized training set for 50 SGD steps, and denote the obtained centralized model as $\theta_{\text{Cen}}$. Then, we use $\theta_0$ as global parameters and distribute it onto $K=10$ clients constructed via the aforementioned split ways. We update $\theta_0$ separately on these 10 clients for 50 steps, and denote the updated models as $\{\theta_{\text{Dec},k}\}_{k =1}^K$. In FedAvg, these updated models will be averaged on the server as the aggregated model, i.e., $\theta_{\text{Agg}}$. We plot the extracted features of $\theta_0$, $\theta_{\text{Cen}}$, and $\theta_{\text{Agg}}$ under various Non-I.I.D. levels. For MNIST, we set the dimension of the final classification layer as 2 and plot the feature scatters. For SVHN, we extract hidden features and then utilize T-SNE~\cite{TSNE} to obtain the 2-dimensional scatters. The figures are plotted in Fig.~\ref{fig:obser} where clusters with different colors represent different classes. The decentralized I.I.D. scenes (i.e., $\overline{C}=10$, $\alpha=10.0$) tend to perform better than centralized training because the former uses $10\times$ training samples (10 clients). However, Non-I.I.D. data (i.e., $\overline{C}=3$, $\alpha=0.1$) experiences performance degradation and the features are less discriminative. In many FL theoretical analyses, local gradient variance among clients is always assumed to be bounded~\cite{DANE,OnTheConverge-ICLR2020,Scaffold}, i.e., $E_k\left[\lVert \nabla_{\theta_k}f_k(\bfx;\theta_k) \rVert^2 \right] \leq \delta, \forall \bfx$. Intuitively, smaller gradient dissimilarity corresponds to better performances and faster convergence. We calculate the gradient variance as $\frac{1}{K}\sum_{k=1}^K \lVert \theta_{\text{Dec},k} - \theta_{\text{Agg}} \rVert^2$. Furthermore, the weight divergence proposed in~\cite{Fed-NonIID-Data} could also reflect the impact of Non-I.I.D. data, i.e., $\frac{\lVert \theta_{\text{Agg}} - \theta_{\text{Cen}} \rVert^2}{\lVert \theta_{\text{Cen}} \rVert^2}$. We calculate these two statistical measures under three Non-I.I.D. levels and plot the bars in Fig.~\ref{fig:obser}, where Non-I.I.D. scenes really lead to larger local gradient variance and weight divergence. These findings conform to previous studies~\cite{Fed-NonIID-Data,FedProx,FedRS}. Additionally and originally, we also investigate the performance gap between Non-I.I.D. and I.I.D. training along with the quality of model initialization, i.e., varying pre-training steps to obtain different $\theta_0$. We plot the performances of $\theta_0$, $\theta_{\text{Cen}}$, and $\theta_{\text{Agg}}$ under various Non-I.I.D. levels in rightmost of Fig.~\ref{fig:obser}. The gap is significantly large when the initialization model is worse, while it reduces a lot with $\theta_0$ becoming better. {\it This observation inspires us that solving the Non-I.I.D. problem in the beginning communication rounds of FL could be more valuable to accelerate training.} To be brief, Non-I.I.D. data lead to performance degradation of FedAvg. 

	\subsection{Federated Ensemble Distillation (FedDF)} \label{sec:feddf}
	FedAvg takes an inductive manner and does not make utilization of the available test data in TFL. FedDF~\cite{FedDF} could use ensemble distillation~\cite{KD} to fine-tune the aggregated model on the unlabeled test data in TFL. Mathematically, instead of simply averaging parameters as done in FedAvg, FedDF takes additional distillation steps to update the global aggregated model as follows:
	\begin{equation}
		\calL_{\text{KL}, j} = \text{KL}\left(\underbrace{\sigma \left( \frac{1}{\vert S_t\vert } \sum_{k\in S_t} f_k(\bfx;\theta_{k}) \right)}_{\text{Distillation Targets}}, \sigma(f(\bfx;\theta_{j-1})) \right), \label{eq:feddf-kl}
	\end{equation}
	\begin{equation}
		\theta_{j} \leftarrow \theta_{j-1} - \eta \nabla_{\theta_{j-1}} E_{\bfx \sim p_{\text{g}}(\bfx)} \left[ \calL_{\text{KL}, j} \right], \label{eq:feddf-update}
	\end{equation}
	where $\theta_j$ is the aggregated model after the $j$th distillation step. $\text{KL}$ denotes KL-divergence usually used in knowledge distillation~\cite{KD}. The used $\bfx$ is originally obtained from a relevant public data set in FedDF, while we could directly sample $\bfx$ from the pre-available test data, i.e. $\sim p_{\text{g}}(\bfx)$, in TFL.
	
	\begin{figure}[htp]
		\centering
		\begin{minipage}{\linewidth}
			\centering
			\includegraphics[width=\linewidth]{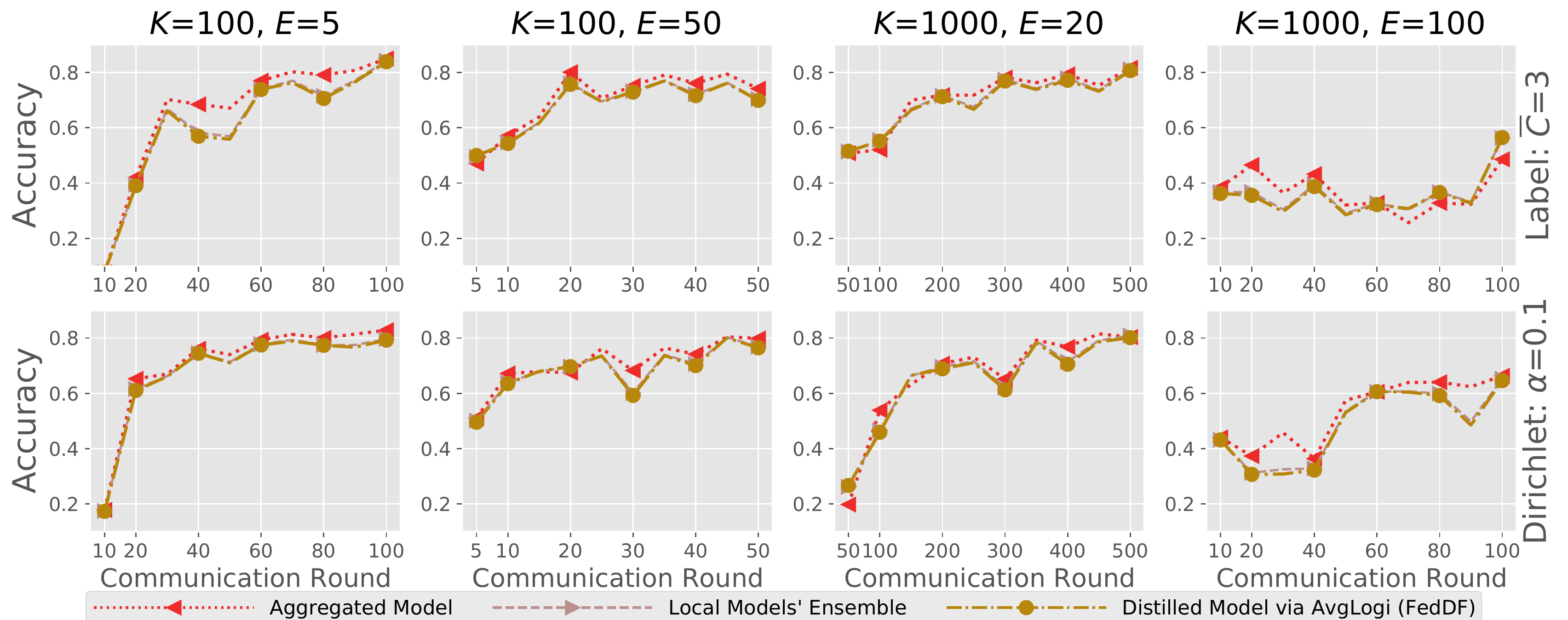}
		\end{minipage}
		\centering
		\caption{\small Performance degradation of FedDF with larger number of clients (e.g., $K=100, 1000$) and lower stochastic participation ratios (e.g., $10\%, 1\%$). Rows show cases split by ``label" and ``dirichlet". $E$ denotes the number of local training epochs.}
		\label{fig:drawback}
	\end{figure}
	
	FedDF significantly depends on the ensemble quality of local models, which is named as {\bf distillation targets} in this paper (Eq.~\ref{eq:feddf-kl}). The verified FL scenes in FedDF are cross-silo ones~\cite{Fed-Advances}, where the number of clients is small (e.g., 20 clients) and clients' participation is stable (e.g., 40\% or 100\% client participation ratio). Furthermore, as declared in FedDF, local clients should undertake more local training steps (e.g., 40 or up to 160 epochs) to obtain ensemble models with enough diversity. These conditions may be too rigorous for some edge devices with unstable communication or limited computation. 
	We consider a larger number of local clients (e.g., 100, 1000) and a smaller client participation ratio (e.g., 10\%, 1\%) in this paper. We run FedDF on decentralized SVHN as an example and plot the results in Fig.~\ref{fig:drawback}. We record test accuracies of the aggregated model, the ensemble of local updated models via ``AvgLogi", and the distilled model obtained via Eq.~\ref{eq:feddf-kl} and Eq.~\ref{eq:feddf-update}. Clearly, the ``AvgLogi" distillation does not improve the aggregated model, and leads to training instability. Therefore, directly applying FedDF to TFL scenes seems to encounter some issues. We attribute the ineffectiveness of FedDF under these scenes to two reasons: {\it varying magnitudes} and {\it improper distillation}. We will detail these in the next section. {\it The essence of FedDF inspires us to propose more effective techniques to refine the inaccurate aggregated model.}
	
	
	\section{Proposed Methods}
	In this section, we introduce our proposed methods. We follow FedDF~\cite{FedDF} and polish it to be broadly applicable to TFL under more settings. Specifically, we propose \underline{M}odel \underline{r}efinery for \underline{T}ransductive \underline{F}ederated learning (MrTF) containing three modules: (1) stabilized teachers; (2) rectified distillation; (3) clustered label refinery. 
	
	\subsection{Stabilized Teachers} \label{sec:stabilize-teacher}
	FedDF~\cite{FedDF} takes ``AvgLogi" to generate the distillation targets, i.e.,
	\begin{equation}
		\overline{q}_{\text{AL}}(y\vert \bfx) = \sigma\left(\sum_k w_k f_k(\bfx;\theta_k)\right), \label{eq:avglogi}
	\end{equation}
	while we consider another one via ``AvgProb" as follows:
	\begin{equation}
		\overline{q}_{\text{AP}}(y\vert \bfx) = \sum_k w_k \sigma\left(f_k(\bfx;\theta_k)\right), \label{eq:avgprob}
	\end{equation}
	where we add weights for each client $w_k \geq 0$ satisfying $\sum_k w_k= 1$, and temporarily omit the client selection process for simplification (i.e., the $\vert S_t\vert $ in Eq.~\ref{eq:feddf-kl}). We calculate the sensitivity of the targets $\overline{q}_{\star,c}$, $c \in [C]$, $\star \in \{\text{AL}, \text{AP}\}$ with respect to the local model parameters $\theta_k$ via calculating the gradients:
	\begin{equation}
		\frac{\partial \overline{q}_{\star,c}}{\partial \theta_k} = w_k J_{\star, c}(y\vert \bfx)\left( \frac{\partial f_{k,c}}{\partial \theta_k} - \sum_{j} J_{\star, j}(y\vert \bfx) \frac{\partial f_{k,j}}{\partial \theta_k} \right), \label{eq:sensitivity}
	\end{equation}
	where $J_{\text{AL}}(y\vert \bfx)=\overline{q}_{\text{AL}}(y\vert \bfx)$, $J_{\text{AP}}(y\vert \bfx)=q_{k}(y\vert \bfx;\theta_k)$. Obviously, the sensitivity is partially determined by the absolute value of the predicted probabilities $J_{\star}(y\vert \bfx)$. {\it This implies that large probabilities make the distillation process sensitive to local models, while moderate prediction results are more stable.} This also consists with some previous distillation research that find tolerant teachers will educate better students~\cite{TolerantKD}. Actually, we find that local updated models could generate ``logits" with varying magnitudes on the same class, making ``AvgLogi" suffer from large variance, and the predicted probabilities vary significantly across classes. We will show observations in experiments (Sect.~\ref{sec:demo}).
	
	To further reduce the ``logits" variance and the sensitivity, we also normalize the ``logits" before calculating probabilities in Eq.~\ref{eq:avgprob} as follows:
	\begin{equation}
		q_k(y\vert \bfx;\theta_k) = \sigma\left(\tau * \frac{f_k(\bfx;\theta_k)}{\text{std}\left(\{f_{k,c}(\bfx;\theta_k)\}_{\bfx \sim p_\text{g}(\bfx), c \in [C]}\right)}\right), \label{eq:avgprob-norm-local}
	\end{equation}
	\begin{equation}
		\overline{q}_{\text{AP}}(y\vert \bfx) = \sum_k w_k q_k(y\vert \bfx;\theta_k), \label{eq:avgprob-normalize}
	\end{equation}
	where $\text{std}(\{\cdot\})$ calculates the standard deviation of a set of values, i.e., all ``logit" values of all classes on all test samples. $\tau$ is the temperature that controls the entropy and we use $\tau=4.0$. {\it This normalization process could generate magnitude-invariant distillation targets among local models, which are more robust to averaging.}
	
	\subsection{Rectified Distillation} \label{sec:rectify-distill}
	From another aspect, the distillation in FedDF aims to optimize:
	\begin{equation}
		\min_{\theta} E_{\bfx\sim p_\text{g}(\bfx)}\left[ -\sum_{c=1}^C \overline{q}(y=c\vert \bfx) \log q_{\text{g}}(y=c\vert \bfx;\theta) \right], \label{eq:distill-loss}
	\end{equation}
	where we use $\overline{q}(y\vert \bfx)=\overline{q}_{\text{AP}}(y\vert \bfx)$ in Eq.~\ref{eq:avgprob-normalize} without any more consideration of ``AvgLogi". This is just the KL-divergence in Eq.~\ref{eq:feddf-kl}. Then, we rewrite the distillation process as:
	\begin{equation}
		\min_{\theta} E_{\bfx}\left[ -\sum_{c=1}^C \left[ \sum_{k \in S} \frac{w_k}{\sum_{j \in S} w_j} q_k(y=c\vert \bfx;\theta_k) \log q_{\text{g}}(y=c\vert \bfx;\theta) \right] \right], \label{eq:distill-loss-rewrite}
	\end{equation}
	where we consider stochastic client participation (i.e., only $\vert S\vert $ clients) resulted from limited or unstable communication. The ideal optimization of $\theta$ should be minimizing $\text{KL}(p_\text{g}(y\vert \bfx), q_\text{g}(y\vert \bfx;\theta))$, $\forall \bfx \sim p_{\text{g}}(x)$. If we could guarantee $\sum_{k \in S} \frac{w_k}{\sum_{j \in S} w_j} q_k(y\vert \bfx;\theta_k)$ approximates $p_\text{g}(y\vert \bfx)$, the distillation process is unbiased and beneficial. This condition could be basically met in TFL if at least one of the following satisfies: (1) the clients' data distributions are the same with the global one, i.e., the I.I.D. case; (2) full or higher client participation in Non-I.I.D. case. The latter one explains why FedDF~\cite{FedDF} is useful in cross-silo FL scenes. However, with a smaller set of participating clients, and supposing only the $k$th client is selected as an extreme case, we actually minimize $\text{KL}(q_k(y\vert \bfx;\theta_k), q_\text{g}(y\vert \bfx;\theta))$. Because $q_k(y\vert \bfx;\theta_k)$ is fitted to $p_k(y\vert \bfx)$ and $p_k(y\vert \bfx) \propto p_k(y)p_k(\bfx\vert y)$, the distillation implicitly biases the global model $\theta$ to the $k$th client's prior distribution $p_k(y)$. Similarly, with a set of clients $S$, the aggregated model will be updated towards $\sum_{k\in S}\frac{w_k}{\sum_{j \in S} w_j}p_k(y)$. {\it In Non-I.I.D. cases, $\sum_{k\in S}\frac{w_k}{\sum_{j \in S} w_j}p_k(y)$ is not guaranteed to cover proper probabilities for all classes and experiences high variance with smaller set of $S$.} For example, we have $C=4$ classes and select $S = \{1,2\}$ with $p_1(y) = [0.5, 0.5, 0.0, 0.0]$ and $p_2(y) = [0.5, 0.0, 0.5, 0.0]$. We use uniform weights. Then the distillation process will be biased towards the distribution $[0.5, 0.25, 0.25, 0.0]$, which brings a negative transfer to the fourth class. We will verify this more in Sect.~\ref{sec:demo}.
	
	We propose two techniques to rectify the distillation targets. The first one is {\it enlarging the ensemble}. The initial global model and the aggregated model in the $t$th round are $\theta_t$ and $\theta_{t+1}$, respectively, while the collected local models are $\{\theta_{t,k}\}_{k\in S_t}$. We use all of these models to generate distillation targets. Considering the aggregated models may perform worse in the beginning, we set a lower weight for them at previous communication rounds and gradually increase the weight. The second technique considers {\it a certain local model could only perform well on a portion of test data}. For example, if a local model only or majorly observes dogs and cats during local training, it could not teach or negatively teach the aggregated model to identify cars. We propose using the predicted entropy to measure how confident is the local model on the predicted sample, i.e., $e_{t,k}(\bfx) = -\sum_{c=1}^C q_k(y=c\vert \bfx;\theta_{t,k})\log q_k(y=c\vert \bfx;\theta_{t,k})$. A smaller entropy corresponds to more confidence, and we let this model contribute more to the distillation process on this sample. Mathematically, the proposed rectified distillation targets are formulated as:
	\begin{eqnarray}
		\nonumber
		\overline{q}_{\text{RAP},t}(y\vert \bfx) &= u_t * \left( \sum_{k \in S_t} \frac{w_{t,k}(\bfx)}{\sum_{j \in S_t} w_{t,j}(\bfx)} q_{k}(y\vert \bfx;\theta_{t,k})  \right)  \\
		&+ \frac{1-u_t}{2} * \underbrace{\left( q_\text{g}(y\vert \bfx;\theta_t) + q_\text{g}(y\vert \bfx;\theta_{t+1}) \right)}_{\text{Self Teaching}}, \label{eq:rap}
	\end{eqnarray}
	where utilizing $\{w_{t,k}(\bfx)\}_{k \sim S_t} = \sigma\left( -1.0 * \{e_{t,k}(\bfx)\}_{k \in S_t} \right)$ can choose appropriate local models for each test sample to generate distillation targets. $q_k(\cdot)$ and $q_{\text{g}}(\cdot)$ are calculated as in Eq.~\ref{eq:avgprob-norm-local}. $u_t$ balances the influence of local and global models. We adjust $u_t$ via $u_t = 0.25 + 0.75 * \left(\frac{1}{\vert S_t\vert }\sum_{k\in S_t}\calL_{t,k}\right) / \log C$. $\calL_{t,k}$ denotes the local cross-entropy loss. With the loss becoming smaller, the aggregated models usually perform better and we enhance their influences. {\it Notably, fusing the initial model $\theta_t$ (i.e., the aggregated model in previous round) and the aggregated model $\theta_{t+1}$ could work as temporal ensembling or self-teaching such as in~\cite{TemporalEnsembling,MeanTeacher,FedPHP}.}
	
	\subsection{Clustered Label Refinery} \label{sec:cluster-refinery}
	The aforementioned two modules separately provide solutions for the problem of varying magnitudes and improper distillation in FedDF~\cite{FedDF}. Only with these two modules, we could already yield better performances compared with FedDF. However, we additionally introduce other techniques to further enhance the stability and quality of the distillation targets. We take advantage of deep clustering~\cite{DeepCluster} to {\it consider feature structural information}. This technique has been verified beneficial in domain adaptation~\cite{SHOT} and transductive few-shot learning~\cite{TransductiveFewShot-ICLR2019}. Formally, we denote the obtained distillation targets as $\overline{q}(y\vert \bfx)=\overline{q}_{\text{RAP}, t}$ (Eq.~\ref{eq:rap}). We extract hidden feature representations via the aggregated global model $\theta_{t+1}$ and denote the features as $\{h(\bfx)\}_{\bfx \sim p_{\text{g}}(\bfx)}$. Then we further improve the distillation targets:
	\begin{equation}
		\bfv_{c} = \frac{E_{\bfx \sim p_{\text{g}}(\bfx)}\left[ \overline{q}_{c}(y\vert \bfx) h(\bfx) \right]}{E_{\bfx \sim p_{\text{g}}(\bfx)}\left[ \overline{q}_{c}(y\vert \bfx) \right]}, \label{eq:refine1}
	\end{equation}
	\begin{equation}
		\overline{q}(y\vert \bfx) = \sigma\left( \{-1.0 * \tau * D_f(h(\bfx), \bfv_c) \}_{c=1}^{C} \right), \label{eq:refine2}
	\end{equation}
	where $D_f(\cdot, \cdot)$ is a distance metric and we use $D_f(\bfx_1, \bfx_2)=1.0 - \frac{\bfx_1^T\bfx_2}{\lVert \bfx_1\rVert \lVert \bfx_2\rVert}$. $\tau$ is the temperature which is also set as $4.0$. The two steps in Eq.~\ref{eq:refine1} and Eq.~\ref{eq:refine2} could be iterated for several steps as done in unsupervised clustering~\cite{DeepCluster}, while we only take one step and it is enough to generate better distillation targets. {\it Notably, the aggregated model could not extract discriminative features in the beginning, thus we omit this process in the first several rounds (e.g., 5).}
	
	\begin{figure}[tbp]
		\includegraphics[width=\linewidth]{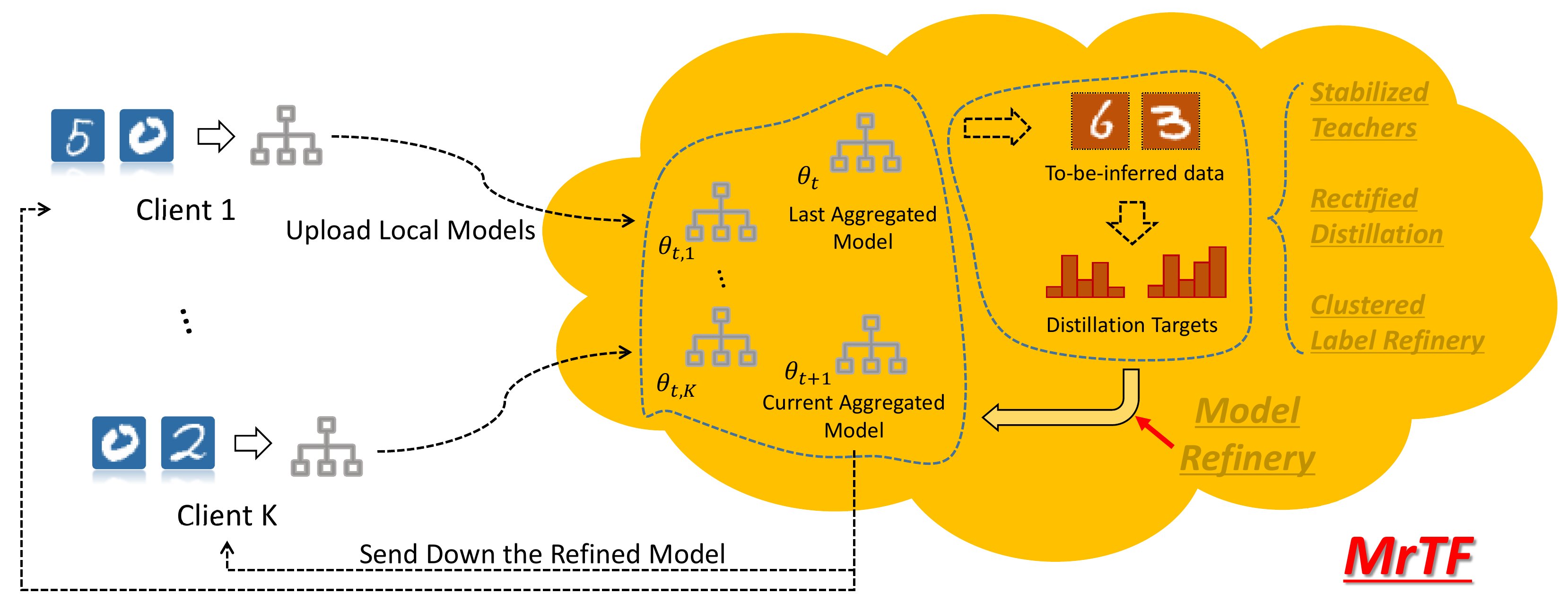}
		\caption{\small The training procedure of the proposed MrTF. The proposed three techniques (i.e., stabilized teachers, rectified distillation, and clustered label refinery) could generate better distillation targets and facilitate the model refinery process.}
		\label{fig:algo}
	\end{figure}
	
	\subsection{MrTF}
	With the three modules, we propose MrTF as follows. During the $t$th communication round, the local procedure is the same as FedAvg~\cite{FedAvg}, while the global procedure takes several steps: (1) collect $\theta_t$, the updated models $\{\theta_{t,k}\}_{k \in S_t}$, and the aggregated model $\theta_{t+1}$; (2) make predictions for the global test set using these models in Eq.~\ref{eq:avgprob-norm-local}; (3) rectify these predicted probabilities in Eq.~\ref{eq:rap}; (4) generate distillation targets via considering feature clusters in Eq.~\ref{eq:refine1} and Eq.~\ref{eq:refine2}; (5) refine the aggregated model $\theta_{t+1}$ on the global test set in Eq.~\ref{eq:feddf-kl} and Eq.~\ref{eq:feddf-update} with the replaced distillation targets. The refined global model is then distributed onto another set of clients for the next round of learning. The procedure of MrTF is illustrated in Fig.~\ref{fig:algo}. The upload and download process is the same as FedAvg. The proposed stabilized teachers, rectified distillation, and clustered label refinery are aimed at generating better distillation targets. The refined model could simultaneously tackle the data heterogeneity challenge across clients and fuse the structural information of the to-be-inferred data.

	\subsection{More Discussion}
	We present more analysis of MrTF from other relevant aspects.
	
	\noindent \textbf{Individual Distillation.} ``AvgProb" in Eq.~\ref{eq:avgprob} could bring another advantage that the distillation could be clearly decomposed into each client, which is more intuitive to analyze. Specifically, the loss in Eq.~\ref{eq:distill-loss-rewrite} could be viewed as $\sum_{k\in S}\frac{w_k}{\sum_{j\in S}w_j} \text{KL}(q_k(y \vert \bfx;\theta_k), q_{\text{g}}(y \vert \bfx;\theta))$, where each client's model individually serves as a teacher. {\it Hence, we expect different teachers transfer different knowledge, i.e., their confident samples, implying the applied weights in Sect.~\ref{sec:rectify-distill} are more rational.}
	
\begin{figure}[tb]
		\centering
		\begin{minipage}{\linewidth}
			\centering
			\includegraphics[width=\linewidth]{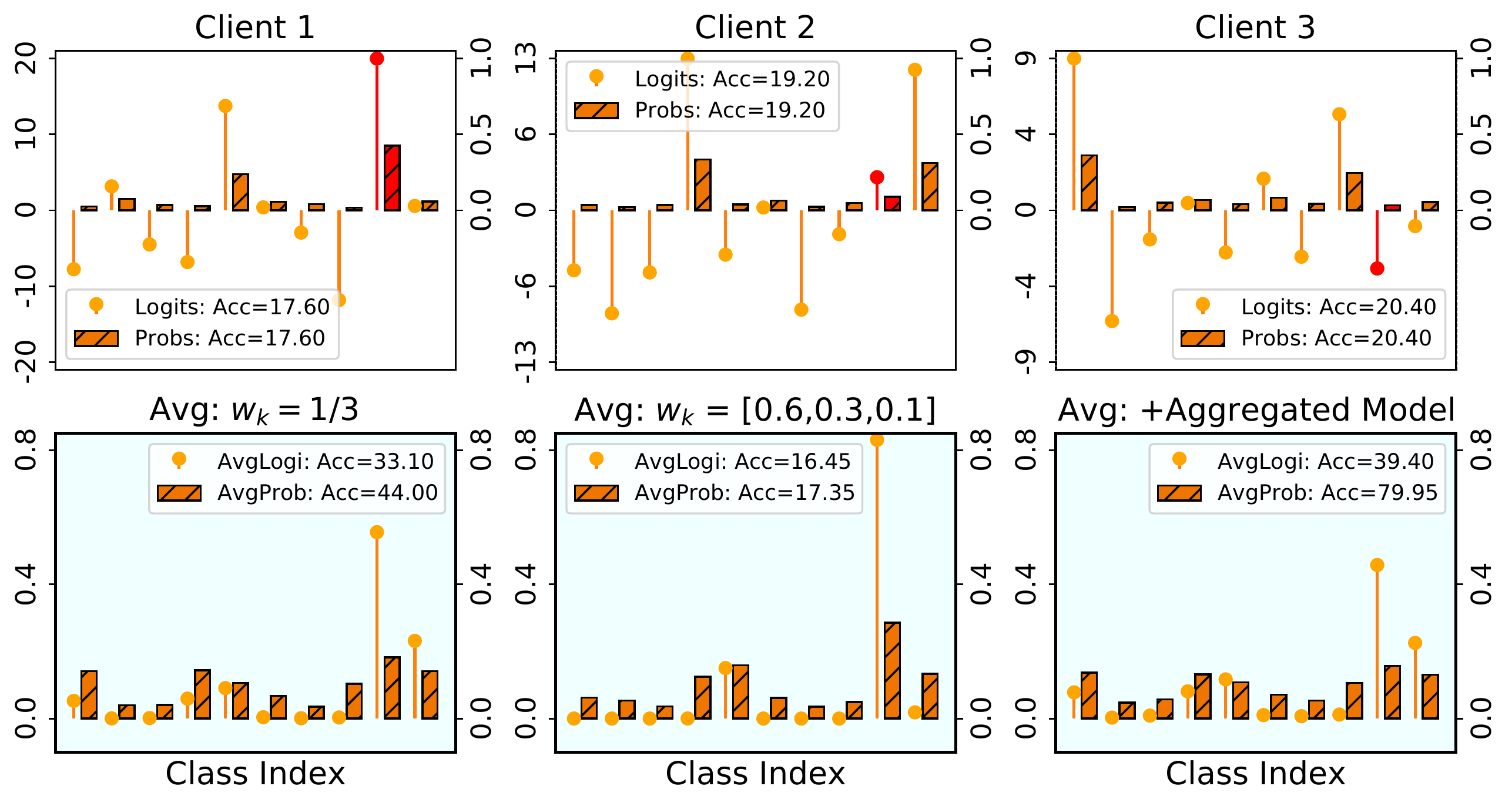}
		\end{minipage}
		\centering
		\caption{\small Comparisons of ``AvgLogi" and ``AvgProb" across three clients on MNIST, each client only observes 2 classes. The top shows the instance-averaged ``logits" and ``probs" on global test set of each local model. The bottom shows the distillation targets generated via: uniform averaging, non-uniform averaging, averaging after adding aggregated models.}
		\label{fig:logi}
	\end{figure}

	\noindent \textbf{Sensitivity to Weights.} We apply weights $w_k(\bfx)$ in Eq.~\ref{eq:rap}, and we could also add a uniform weight $w_k=1/K$. If we do not use ``AvgProb", directly utilizing ``AvgLogi" has been verified sensitive to different weighting, shown in Fig.~\ref{fig:logi}. Theoretically, the sensitivity of $\overline{q}_{\text{AL}}$ with respect to $w_k$ is relevant to the absolute value of $f_k(\bfx;\theta_k)$ (Eq.~\ref{eq:avglogi}), while in $\overline{q}_{\text{AP}}$, it is relevant to $\sigma(f_k(x;\theta_k)) \in [0, 1]$. Obviously, the latter is more robust to the applied weights. This paves the foundation for adding two-level weights in rectified distillation (Sect.~\ref{sec:rectify-distill}).
	
	\noindent \textbf{Self Teaching.} In the module of rectified distillation (Sect.~\ref{sec:rectify-distill}), we add global aggregated models into the ensemble. We could decompose Eq.~\ref{eq:rap} into three parts: (1) the first is distilling local models' ability to the aggregated model; (2) the second is like $\text{KD}(\theta_t, \theta_{t+1})$, which utilizes historical prediction to supervise the current learning; (3) the third part is $\text{KD}(\theta_{t+1}, \theta_{t+1})$, which is similar to self-teaching. $\text{KD}(\cdot, \cdot)$ denotes the knowledge distillation process.
	
	\begin{table*}
		\centering
		\caption{{\small Statistics of utilized datasets. We take $K=100$ for an example.}}
		\label{tab:data}
		{
			\begin{tabular}{@{}l|c|c|c|c|c|c@{}}
				\toprule
				& $N$ & $M$ & $C$ & $K$ & $\overline{N_k}$ & $\overline{C_k}$ \\
				\midrule
				MNIST~\cite{mnist} & 55k & 10k & 10 & 100 & 550 & 3.0 $\vert$ 3.6 \\
				MNISTm~~\cite{Mnistm-DaNN} & 55k & 10k & 10 & 100 & 550 & 3.0 $\vert$ 3.6 \\
				SVHN~\cite{Svhn} & 73k & 26k & 10 & 100 & 730 & 3.0 $\vert$ 3.9 \\
				CIFAR10~\cite{cifar} & 50k & 10k & 10 & 100 & 500 & 3.0 $\vert$ 3.6 \\
				\hline
				CIFAR100~\cite{cifar} & 50k & 10k & 100 & 100 & 500 & 30 $\vert$ 17.5 \\
				FeMnist~\cite{LEAF} & 85k & 16k & 62 & 359 & 236 & 15.3 $\vert$ 14.4 \\
				Shakespeare~\cite{LEAF} & 437k & 84k & 80 & 112 & 3.9k & 28.7 $\vert$ 29.2 \\
				\botrule
			\end{tabular}
		}
	\end{table*}

	\section{Experiments}
	We use datasets from: (a) digits recognition: MNIST~\cite{mnist}, MNISTm~\cite{Mnistm-DaNN}, SVHN~\cite{Svhn}; (b) image classification: CIFAR10/100~\cite{cifar}, recommended by FedML~\cite{FedML}; (c) FeMnist and Shakespeare, recommended by LEAF~\cite{LEAF}. Datasets in (a) and (b) are commonly utilized as benchmarks in centralized training. In our work, we split the corresponding training set onto $K$ clients according to ``split by label" with different $\overline{C}$ or ``split by dirichlet" with different $\alpha$. Smaller $\overline{C}$ and $\alpha$ lead to more Non-I.I.D. scenes, i.e., clients' data distributions differ a lot. Benchmarks in (c) provide a user list, and we construct Non-I.I.D. FL scenes via taking each user as an individual client. Specifically, Shakespeare is a dataset built from the Complete Works of William Shakespeare, which is originally used in FedAvg~\cite{FedAvg}. It is constructed by viewing each speaking role in each play as a different device, and the target is to predict the next character based on the previous characters. FeMnist is a task to classify the mixture of digits and characters, where data from each writer is considered as a client. These two benchmarks contain amounts of training samples and we only select $10\%$ data for training. We list statistics of these benchmarks in Tab.~\ref{tab:data} including: (1) the total amount of training samples of all clients ($N$); (2) the total number of test samples on the server ($M$); (3) the number of classes ($C$); (4) the number of clients ($K$); (5) the number of training samples of each client on average ($\overline{N_k}$); (6) the number of observed classes (i.e., at least 5 training samples) of each client on average when split by label ($\overline{C}=3$) or dirichlet ($\alpha=0.1$), denoted as $\overline{C_k}$ separated by "$\vert$". 
	
	For different datasets, we use corresponding deep neural networks, including: (1) MLPNet for MNIST with three layers, the hidden size of each hidden layer is 1024, and the last layer's size is 2 for visualization in Fig.~\ref{fig:obser} and 128 for performance comparisons; (2) LeNet~\cite{mnist} for MNISTm; (3) ConvNet for SVHN as used in FedAvg~\cite{FedAvg}, we use T-SNE~\cite{TSNE} for visualization in Fig.~\ref{fig:obser}; (4) VGG8~\cite{VGG} for CIFAR10/100 with 5 convolution layers and 3 fully-connected layers; (5) ResNet8/20~\cite{ResNet} for CIFAR100; (6) FeCNN for FeMnist as used in LEAF~\cite{LEAF}; (7) CharLSTM for Shakespeare as used in FedAvg~\cite{FedAvg}. For our proposed MrTF, we extract features for further label refinery as introduced in Sect.~\ref{sec:cluster-refinery}. For MLPNet and CharLSTM, we utilize the last hidden layer's output as features; for convolution networks, we use the flattened convolution features.
	
	In TFL, the number of clients $K$, the client participation ratio $R$, the split parameters $\overline{C}$ and $\alpha$ determine a FL scene. Usually, $K$ is large in FL, and $R$ could be small due to limited or unstable communication. $\overline{C}$ and $\alpha$ are introduced to split the centralized training data for simulating a decentralized setting. We investigate $K=100, 1000$, $R=10\%,1\%$ in our experiments. We also investigate several data split ways, e.g., $\overline{C}=5,3$ for $C=10$, $\alpha=1.0,0.1$. Smaller $\overline{C}$ or $\alpha$ corresponds to more Non-I.I.D. scenes. Other important hyper-parameters include the number of global communication rounds $T$ and the local training epochs $E$. We also study our method on various settings of $T$ and $E$. We use SGD with a momentum of 0.9 as the local optimizer. For digits recognition scenes, we vary learning rate in $\{0.1, 0.05, 0.01\}$ and report the best one for comparison; for CIFAR scenes, we vary learning rate in $\{0.05, 0.03, 0.01\}$; for FeMnist, we use $0.004$; for Shakespeare, we use $1.47$. For digits and CIFAR scenes, we use a batch size of 64; for FeMnist and Shakespeare, we use 10. We use Adam with a learning rate $0.0003$ as the global optimizer in FedDF and MrTF (Ours) and take 500 distillation steps.
	
	\subsection{Demo Analysis} \label{sec:demo}
	We first verify the success of the first two modules in MrTF, which are proposed to tackle the varying magnitudes and improper distillation drawbacks in FedDF~\cite{FedDF}.
	We experiment on MNIST with three clients and each client could only observe two classes. We init a global model and distribute it to the three clients. After local training, we use these three local models to predict on the global test set, recording the accuracy and each instance's ``logits" and ``probs". We average the class ``logits" or ``probs" across test samples for better presentation. Because the global test set is uniformly distributed across 10 classes, and {\it we expect the average results of both ``logits" and ``probs" are also uniform}. The results are shown in Fig.~\ref{fig:logi}.
	
	First, the top three figures show the results of each local model. The accuracies are low, i.e., 17.6\%, 19.2\%, 20.4\%. The reasons are intuitive: they are trained with only 2 classes, while the global test set contains 10 classes. The ``logits" across clients vary greatly, with the largest ranging from 9.0 to 20.0, while the corresponding ``probs" are limited to $[0,0.5]$. If we uniformly ($w_k=1/K$) average ``logits" and ``probs" of three local models for each test instance as done in Eq.~\ref{eq:avglogi} and Eq.~\ref{eq:avgprob}, the results are shown at the bottom left of Fig.~\ref{fig:logi}. Because the 9th class (showed in red) generally has large ``logits" (i.e., around 20.0) predicted by the first local model, it dominates the $\sigma(\cdot)$ operation and makes ``AvgLogi" output much higher probabilities on the 9th class. However, using ``AvgProb" leads to smoother class probabilities and the test accuracy improves from 33.1\% to 44.0\%. If we apply $w_1=0.6$,$w_2=0.3$,$w_3=0.1$ for averaging, the results of ``AvgLogi" are worse as shown in the bottom middle of Fig.~\ref{fig:logi} (the 9th stem is higher). However, ``AvgProb" performs more stably and the class probabilities are more uniform. {\it These observations show that replacing ``AvgLogi" with ``AvgProb" could really mitigate the problem of the varying magnitude, leading to moderate teachers and better ensemble performances.}
	
	From another aspect, because these three clients only observe at most 6 classes in total, some unseen classes' ``logits" will be inaccurate. Illustrated in Fig.~\ref{fig:logi}, some classes' probabilities become zero. That is, the stochastic client participation will lead to inaccurate distillation targets, and directly using ``AvgLogi" or ``AvgProb" for distillation is improper. Instead, we fuse the global aggregated models and rectify the probabilities as done in Sect.~\ref{sec:rectify-distill}. Then the results shown at the bottom right of Fig.~\ref{fig:logi} are better. That is, the probabilities of ``AvgLogi" and ``AvgProb" become more smooth and the test accuracies are improved to 39.4\% and 79.9\%, respectively. All observations verify the rationality and effectiveness of our solutions in Sect.~\ref{sec:stabilize-teacher} and Sect.~\ref{sec:rectify-distill}.
	
	\begin{figure}[t]
		\centering
		\begin{minipage}{\linewidth}
			\centering
			\includegraphics[width=\linewidth]{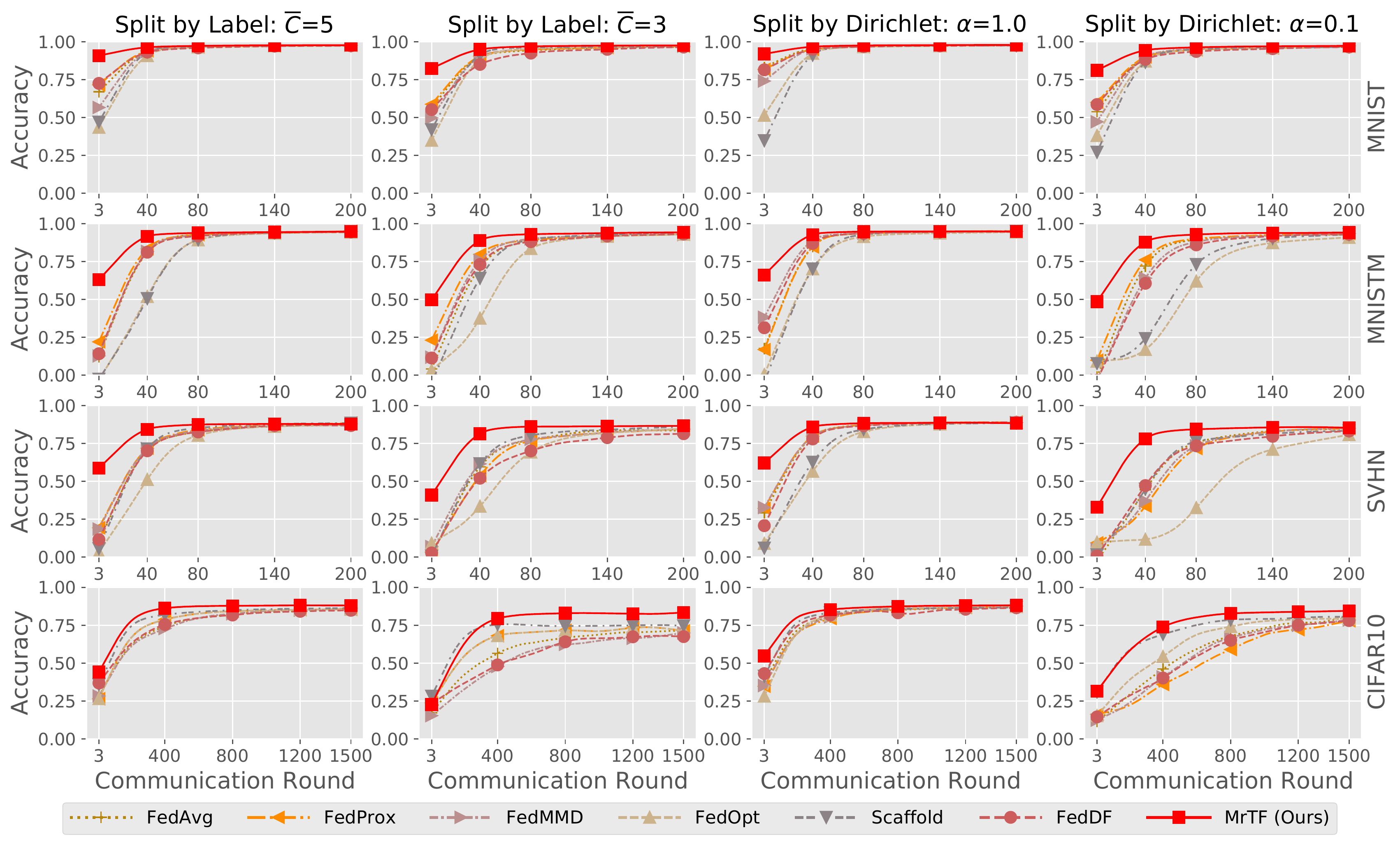}
		\end{minipage}
		\centering
		\caption{\small Performance comparisons on several FL scenes. Row shows each dataset and column shows each data split way.}
		\label{fig:compare}
	\end{figure}
	
	\begin{figure}[t]
		\centering
		\begin{minipage}{\linewidth}
			\centering
			\includegraphics[width=\linewidth]{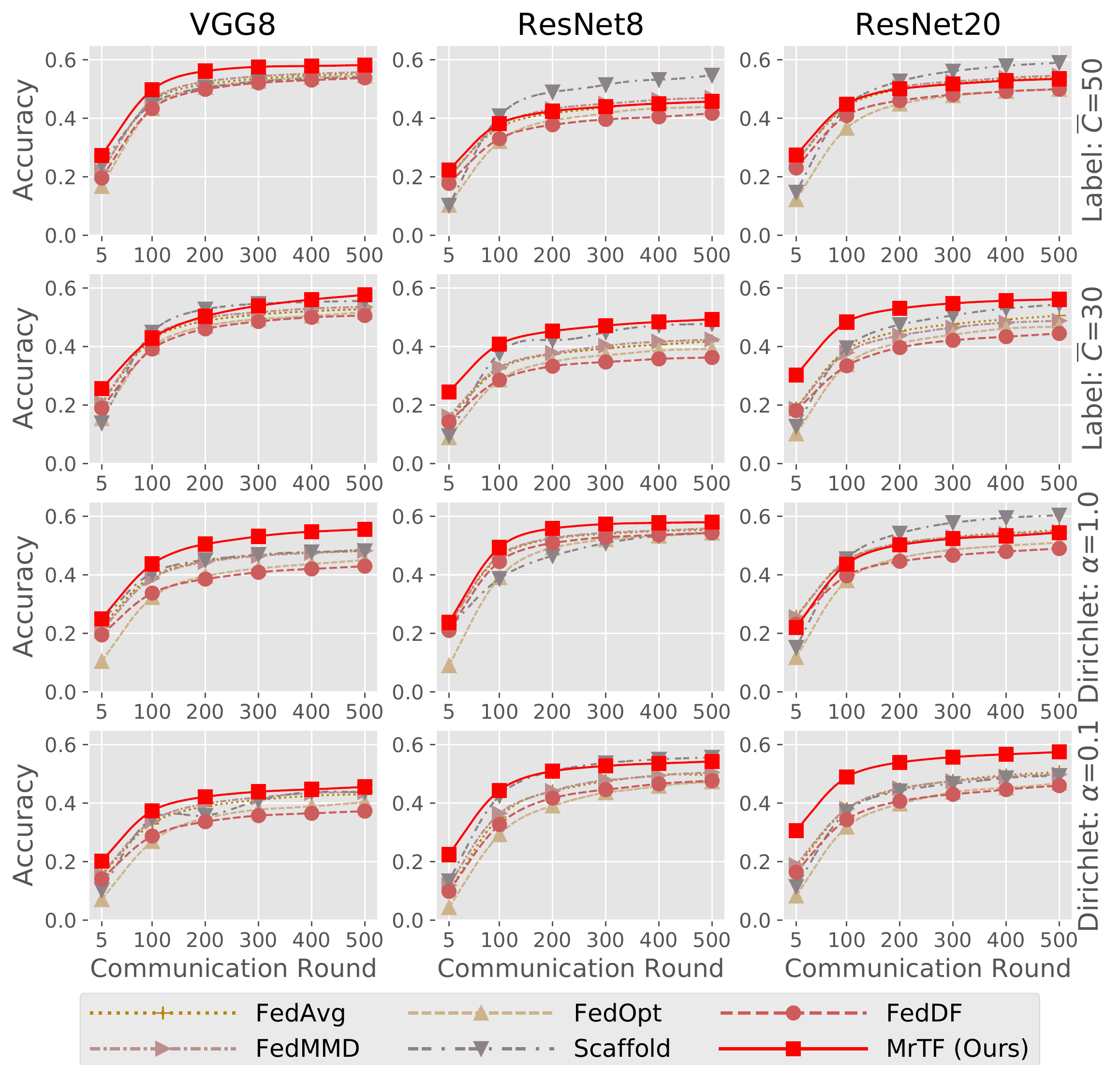}
		\end{minipage}
		\centering
		\caption{\small Performance comparisons on CIFAR100 based on VGG8 and ResNet8/20. Row shows FL scene with different data split ways.}
		\label{fig:compare-cifar100}
	\end{figure}
	
	\begin{figure}[htp]
		\centering
		\begin{minipage}{\linewidth}
			\centering
			\includegraphics[width=\linewidth]{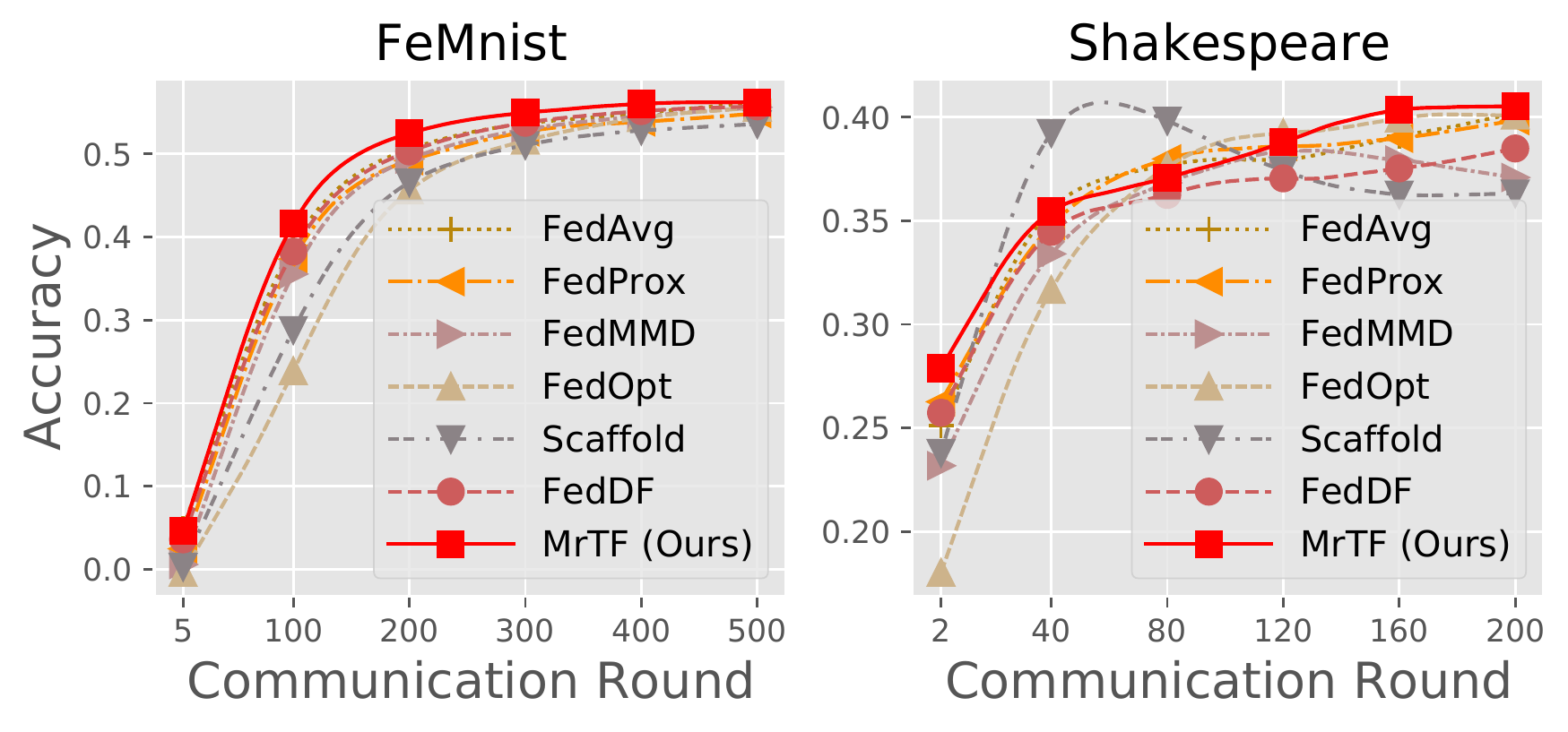}
		\end{minipage}
		\centering
		\caption{\small Performance comparisons on LEAF benchmarks.}
		\label{fig:compare-other}
	\end{figure}

	\subsection{Performance Comparisons} \label{sec:compare}
	We compare MrTF with FedAvg~\cite{FedAvg}, FedProx~\cite{FedProx}, FedMMD~\cite{FedMMD}, FedOpt~\cite{FedOpt}, Scaffold~\cite{Scaffold}, and FedDF~\cite{FedDF}. The first five algorithms do not access the global test set during training, while they utilize various techniques to solve Non-I.I.D. problems. FedDF utilizes ``AvgLogi" to refine the aggregated model, which is the most similar to ours. Details of these algorithms are presented as follows.
	\begin{itemize}
		\item \textbf{FedAvg~\cite{FedAvg}}: the most standard FL algorithm that utilizes parameter averaging for model aggregation.
		\item \textbf{FedProx~\cite{FedProx}}: introduces a proximal term during local procedures to constrain the model parameters' update.
		\item \textbf{FedMMD~\cite{FedMMD}}: introduces the discrepancy minimizing optimization (i.e., MMD) in local procedures and regularizes the local model not diverge a lot from the global model too much.
		\item \textbf{FedOpt~\cite{FedOpt}}: updates the global model via momentum or adaptive optimization techniques to stabilize the global model's update.
		\item \textbf{Scaffold~\cite{Scaffold}}: points out the local update will diverge from the global direction and utilizes control variates to reduce local gradient variance.
		\item \textbf{FedDF~\cite{FedDF}}: uses local models' ensemble, i.e., ``AvgLogi", to finetune the global model on a relevant public data set.
	\end{itemize}
	
	\noindent \textbf{Part \uppercase\expandafter{\romannumeral1}} We first study on MNIST, MNISTm, SVHN, and CIFAR10, which have 10 classes to identify. We construct four FL scenes for each dataset via ``split by label" with $\overline{C}\in \{5,3\}$ and ``split by dirichlet" with $\alpha \in \{1.0, 0.1\}$. We take $K=100$ clients and only select $R=10\%$ clients in each communication round. We update $E=3$ epochs for each client during local procedures and take $T=200$, $1500$ communication rounds for digits and CIFAR scenes, respectively. The results are shown in Fig.~\ref{fig:compare}, where MrTF converges faster and performs better on all scenes. First, MrTF could surpass other methods by a large margin especially in the beginning, verifying that learning from local models' ensemble significantly helps and lays the foundation for subsequent improvements. This conforms to the observation in Sect.~\ref{sec:Non-I.I.D.}. Some compared algorithms could only improve FL on certain scenes. For example, Scaffold performs better than others on CIFAR10, while worse on other datasets.	
	
	\noindent \textbf{Part \uppercase\expandafter{\romannumeral2}} Then, we vary the utilized networks and compare the performances on CIFAR100 using VGG8\cite{VGG} and ResNet8/20~\cite{ResNet}. We take $K=100$ clients and $R=10\%$. The results are shown in Fig.~\ref{fig:compare-cifar100}. For each scene, we run 500 communication rounds and each client takes on $E=20$ epochs. In some cases, MrTF performs worse than Scaffold, attributed to the control variates used in Scaffold. However, MrTF could obtain better results on most of the cases, especially in more Non-I.I.D. scenes, i.e., $\overline{C}=30$ and $\alpha=0.1$ (the 2nd and 4th row in Fig.~\ref{fig:compare-cifar100}). Because we have 100 classes, the possibility that participating clients cannot cover all classes greatly increases, making FedDF ineffective.
	
	\noindent \textbf{Part \uppercase\expandafter{\romannumeral3}} We also investigate our method on LEAF~\cite{LEAF} benchmarks, i.e., FeMnist and Shakespeare. These two benchmarks are split by users, where the distribution skew dominates the Non-I.I.D.~\cite{Fed-Advances} problem. We show the results in Fig.~\ref{fig:compare-other}. Our proposed MrTF could still show effectiveness towards other methods. Although Scaffold could achieve faster convergence in the beginning, the performance degrades a lot with a larger communication round. The training instability limits the application of Scaffold to TFL. Compared with this, our proposed MrTF could achieve the Scaffold's best performance and is more stable.
	
	\begin{figure}[htp]
		\centering
		\begin{minipage}{\linewidth}
			\centering
			\includegraphics[width=\linewidth]{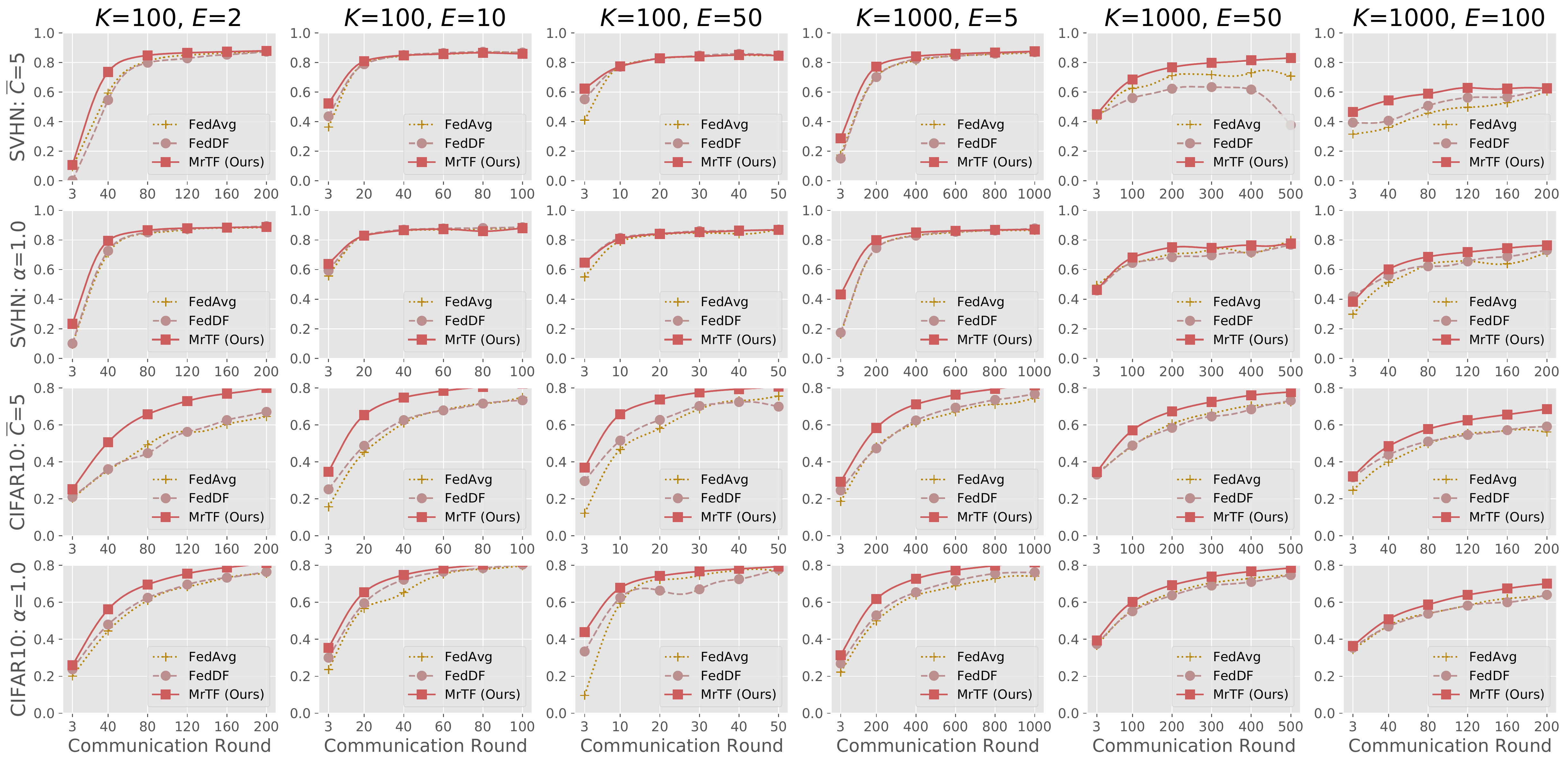}
		\end{minipage}
		\centering
		\caption{\small Comparison results with FedAvg and FedDF on more FL scenes. Row shows dataset and corresponding split strategy, and column shows the number of clients $K$ and the number of local training epochs $E$.}
		\label{fig:scenes}
	\end{figure}
	
	\noindent \textbf{Part \uppercase\expandafter{\romannumeral4}} We then majorly compare with FedAvg and FedDF on various scenes. Specifically, we vary $K \in \{100, 1000\}$ to investigate large amounts of clients. For $K=100$, we take $R = 10\%$ to select only 10 clients in each round and each client updates $E \in \{2, 10, 50\}$ local epochs; for $K=1000$, we use $R=1\%$ and $E\in \{5, 50, 100\}$. We experiment on SVHN and CIFAR10 under two split strategies, i.e., ``split by label" with $\overline{C}=5$ and ``split by dirichlet" with $\alpha=1.0$. The results are plotted in Fig.~\ref{fig:scenes}. MrTF can basically surpass FedAvg and FedDF on all scenes. Additionally, MrTF behaves more stably even with larger number of clients (e.g., $K=1000$), contributed to the stabilized teachers and rectified distillation.
	
	\begin{figure*}[htp]
		\centering
		\begin{minipage}{\linewidth}
			\centering
			\includegraphics[width=\linewidth]{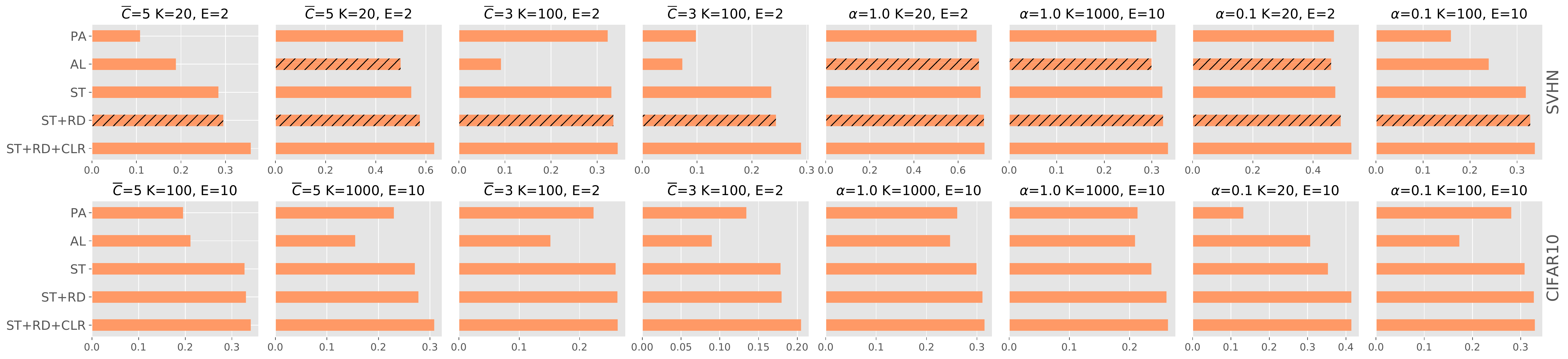}
		\end{minipage}
		\centering
		\caption{\small Ablation studies of the modules in MrTF. Each row shows a data set. Each column shows the split strategy ($\overline{C} \in [5,3]$, $\alpha \in [1.0, 0.1]$) and corresponding $K$, $E$. The five bars refer to test accuracy of: (1) PA (parameter averaging in FedAvg); (2) AL (``AvgLogi" in FedDF); (3) ST (Stabilized Teachers, Sect.~\ref{sec:stabilize-teacher}); (4) ST + RD (Rectified Distillation, Sect.~\ref{sec:rectify-distill}); (5) ST + RD + CLR (Clustered Label Refinery, Sect.~\ref{sec:cluster-refinery}) (MrTF).}
		\label{fig:ablation}
	\end{figure*}
	
	\subsection{Ablation Studies} \label{sec:ablation}
	Our proposed MrTF contains three modules: (1) we use ``AvgProb" in Eq.~\ref{eq:avgprob-normalize} instead of ``AvgLogi" to obtain stabilized teachers; (2) we fuse aggregated models into local models and apply two-level weights for rectified distillation; (3) we additionally take clustering techniques to refine the distillation targets.  We incrementally add these modules for ablation studies. We denote these three components as ``ST", ``RD", and ``CLR". Correspondingly, we compare performances of: (1) simple parameter averaging without any refinery (i.e., FedAvg); (2) averaging logits (i.e., ``AvgLogi" in FedDF); (3) ST; (4) ST + RD; (5) ST + RD + CLR (i.e., proposed MrTF). We compare them on SVHN and CIFAR10 under various FL scenes, and the results could be found in Fig.~\ref{fig:ablation}. For each scene, we run 50 communication rounds. ``AvgLogi" is not stable and sometimes surpasses parameter averaging while sometimes does not. Only using the ``AvgProb" in Eq.~\ref{eq:avgprob-normalize} (i.e., ST) could already yield notable performances, while fusing RD and CLR could lead to higher results.
	
	\subsection{More Studies: Cross-Domain TFL} \label{sec:cross-tfl}
	In some cases, although a party could collaborate with other parties to help infer the handy unlabeled data via the proposed TFL framework, the distribution of the unlabeled data may also be heterogeneous from others. We call this case cross-domain TFL, which is similar to the scene studied in~\cite{FADA,KD3A}. These works only consider several heterogeneous domains (e.g., 5), which are more similar to domain adaptation under privacy protection~\cite{DAN,SHOT}. That is, they do not consider some other challenges in our work, i.e., stochastic client participation, low-shot training samples, class imbalance, etc. In cross-domain TFL, we have to simultaneously tackle these challenges aside from Non-I.I.D. data and cross-domain knowledge transfer. We preliminarily apply MrTF to this scene. Specifically, we split SVHN (MNISTm) data across $K=100$ clients with $\alpha \in \{1.0, 0.1\}$. The server aims to make predictions for MNISTm (SVHN). In each round, we select 10 clients and each client takes on 5 epochs. We run 200 communication rounds and report the final accuracies averaged by 5 independent experiments. We compare with FedAvg and FedDF. Results are listed in Tab~\ref{tab:cross-domain}. MrTF could still surpass FedAvg and FedDF by a significant margin even in cross-domain TFL. However, the overall cross-domain transfer performance is still lower compared with in-domain learning, which means that more advanced domain adaptaion~\cite{DAN} techniques should be considered for cross-domain TFL.
	
	\begin{table*}
		\centering
		\caption{{\small Performance comparisons in cross-domain TFL.}}
		\label{tab:cross-domain}
		{
			\begin{tabular}{@{}l|c|c|c|c@{}}
				\toprule
				& \multicolumn{2}{c|}{SVHN$\rightarrow$MNISTm} & \multicolumn{2}{c}{MNISTm$\rightarrow$SVHN} \\
				& $\alpha$=1.0 & $\alpha$=0.1 & $\alpha$=1.0 & $\alpha$=0.1 \\
				\midrule
				FedAvg~\cite{FedAvg} & 40.26 & 34.52 & 32.09 & 31.06 \\
				FedDF~\cite{FedDF} & 41.90 & 35.91 & 33.06 & 28.84 \\
				MrTF (Ours) & \bf{44.78} & \bf{41.70} & \bf{35.32} & \bf{34.55} \\
				\botrule
			\end{tabular}
		}
	\end{table*}
	
	\subsection{More Studies: Privacy Protection} \label{sec:dp}
	FedAvg could only provide basic privacy protection for users, while some advanced attacks could still break privacy via inverting local gradients~\cite{DLG,InvGradFL}. Hence, techniques such as differential privacy~\cite{DeepDP} should be considered for stricter privacy protections. To guarantee $(\epsilon, \delta)$-DP in FL, gradient clipping is applied to local model updates, and gaussian noises $\mathcal{N}(0, \sigma^2)$ are added before being sent to the server. With added noise, the aggregated model will be more inaccurate. However, we expect our model refinery process could mitigate the performance degradation. We experiment on CIFAR10 with $\alpha=1.0, K=100, E=5, T=200$. We use VGG8 and add noise $\sigma \in \{0.0, 0.001, 0.01, 0.1\}$. We report the results of FedAvg and MrTF in Tab.~\ref{tab:dp}. With higher noise, FedAvg's performance degrades seriously while MrTF could maintain a better prediction.
	
	\begin{table*}
		\centering
		\caption{{\small Performances when adding differential privacy.}}
		\label{tab:dp}
		{
			\begin{tabular}{@{}l|c|c|c|c@{}}
				\toprule
				& $\sigma$=0.0 & $\sigma$=0.001 & $\sigma$=0.01 & $\sigma$=0.1 \\
				\midrule
				FedAvg~\cite{FedAvg} & 79.46 & 76.98 & 69.08 & 38.27 \\
				MrTF (Ours) & \bf{82.31} & \bf{81.76} & \bf{76.54} & \bf{47.11} \\
				\botrule
			\end{tabular}
		}
	\end{table*}

	\subsection{Limitations and Future Work}
	Our proposed MrTF is a novel and practical solution to the introduced real-world scenario in that a newly-established pilot project needs to build a machine-learning model with the help of other isolated parties. However, MrTF does not consider the existing models of these parties and trains local models from scratch, making the convergence slower. Utilizing the available pre-trained models and accelerating the training process may be interesting for future work.
	
	\section{Conclusion}
	We consider transductive federated learning (TFL), where the server owns to-be-referred data while the training data are distributed across other parties. We in-depth analyze some existing FL works and point out their drawbacks. As an alternative, we propose MrTF with three modules, i.e., stabilized teachers, rectified distillation, and clustered label refinery, to refine the global aggregated model and make predictions in a transductive manner. Our proposed method shows superiorities towards compared methods on various investigated scenes.

	\backmatter
	
	\bmhead{Acknowledgments}
	This work is partially supported by the National Natural Science Foundation of China (Grant No. 61921006, 62006118, 62276131), the National Key RD Program of China (Grant No. 2022YFF0712100) and the Fundamental Research Funds for the Central Universities (NO.NJ2022028, No.30922010317). Thanks to Huawei Noah’s Ark Lab NetMIND Research Team.

	
	\bibliographystyle{sn-mathphys}
	\bibliography{manuscript}
	
	
\end{document}